\documentclass[10pt,twocolumn,letterpaper]{article}

\usepackage[pagenumbers]{cvpr} 

\usepackage{amssymb}
\usepackage[utf8]{inputenc}         
\usepackage[T1]{fontenc}            
\usepackage{url}                    
\usepackage{booktabs}               
\usepackage{amsfonts}               
\usepackage{nicefrac}               
\usepackage{microtype}              
\usepackage{xcolor}                 
\usepackage{graphicx}               
\usepackage{amsmath}                
\usepackage{multirow}               
\usepackage{array}                  
\usepackage{comment}                

\DeclareMathOperator*{\argmin}{arg\,min}

\definecolor{cvprblue}{rgb}{0.21,0.49,0.74}
\usepackage[pagebackref,breaklinks,colorlinks,citecolor=cvprblue]{hyperref}

\def\paperID{17250} 
\def\confName{CVPR}
\def\confYear{2024}

\begin{document}

\title{TRIGS: Trojan Identification from Gradient-based Signatures\thanks{This preprint has not undergone peer review or any post-submission improvements or corrections. The Version of Record of this contribution is published in ICPR 2024, and is available online at \url{https://doi.org/10.1007/978-3-031-78122-3_23}.}}

\author{
  Mohamed E.~Hussein \\
  USC Information Sciences Institute\\
  Arlington, VA 22203 \\
  {\tt\small mehussein@isi.edu}
  \and
  Sudharshan Subramaniam Janakiraman \\
  USC Information Sciences Institute \\
  Marina del Rey, CA 90292 \\
  {\tt\small ss20785@usc.edu}
  \and
  Wael AbdAlmageed \\
  Clemson University \\
  Electrical and Computer Engineering Department \\
  Riggs Hall, Clemson, SC 29634, USA \\
    {\tt\small wabdalm@clemson.edu}
}
\maketitle

\begin{abstract}
Training machine learning models can be very expensive or even unaffordable. This may be, for example, due to data limitations (unavailability or being too large), or computational power limitations. Therefore, it is a common practice to rely on open-source pre-trained models whenever possible. However, this practice is alarming from a security perspective. Pre-trained models can be infected with Trojan attacks, in which the attacker embeds a trigger in the model such that the model's behavior can be controlled by the attacker when the trigger is present in the input. In this paper, we present a novel method for detecting Trojan models. Our method creates a signature for a model based on activation optimization. A classifier is then trained to detect a Trojan model given its signature. We call our method TRIGS for TRojan Identification from Gradient-based Signatures. TRIGS achieves state-of-the-art performance on two public datasets of convolutional models. Additionally, we introduce a new challenging dataset of ImageNet models based on the vision transformer architecture. TRIGS delivers the best performance on the new dataset, surpassing the baseline methods by a large margin. Our experiments also show that TRIGS requires only a small amount of clean samples to achieve good performance, and works reasonably well even if the defender does not have prior knowledge about the attacker's model architecture. Our code and data can be accessed through this page \url{github.com/vimal-isi-edu/trigs}.
\end{abstract}

\section{Introduction}

\begin{figure}[ht]
    \centering
    \includegraphics[width=\linewidth]{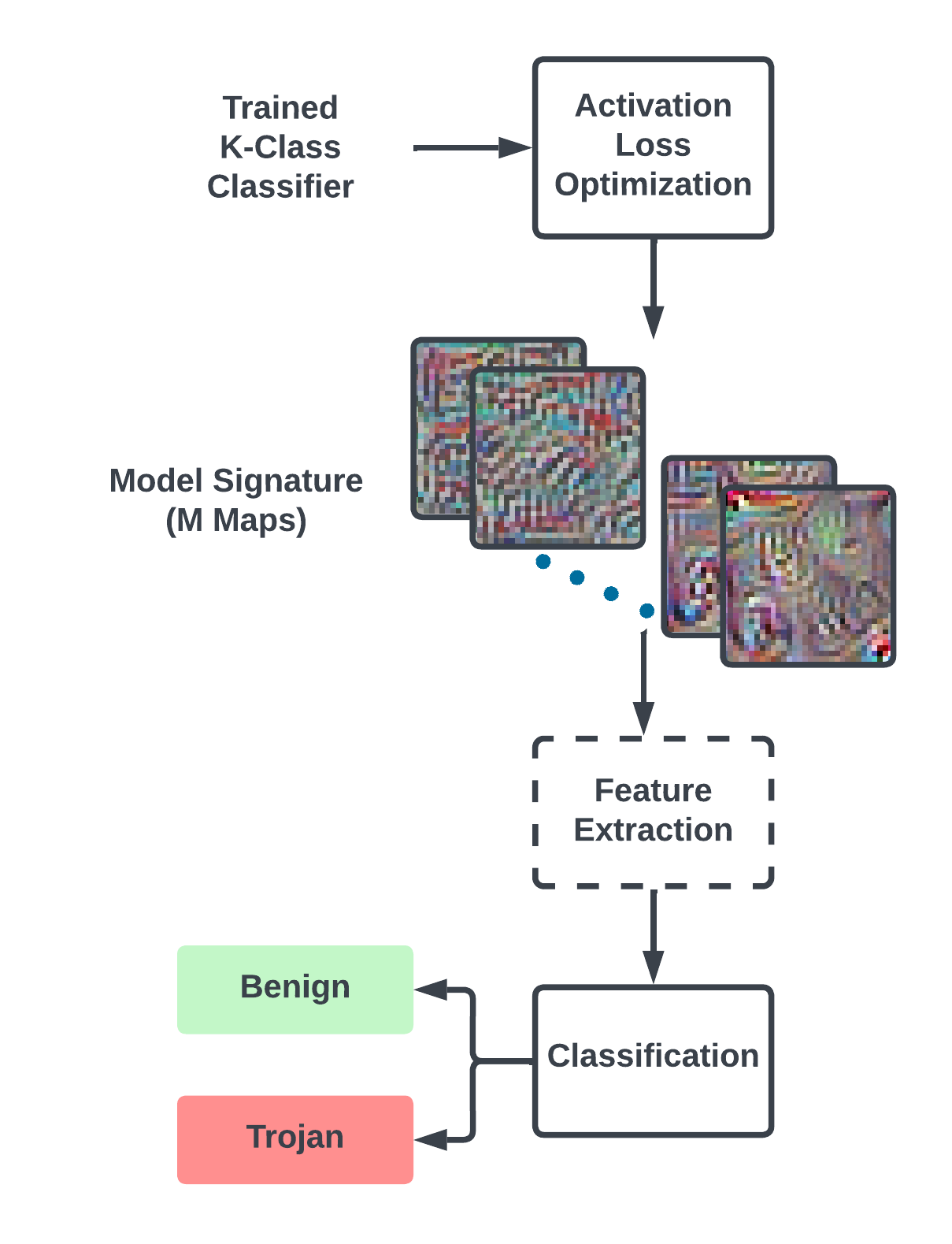}
    \caption{Proposed framework for Trojan model detection. Given a $K$-class classifier, $M$ loss functions are optimized by adapting the input to the model. The resulting images constitute the signature for the model, which is used by a downstream classifier to tell if the model is Trojan or benign, after an optional feature extraction step.}
    \label{fig:framework}
\end{figure}

Machine learning has made great progress since the introduction of deep learning. However, the training of deep models remains more of an art than science. It requires a lot of trial and error and parameter fine-tuning. All this incurs a significant computational cost and energy footprint. More importantly, high-performing models are trained on huge amounts of data, a process that can only be afforded by a few organizations. As a result, researchers and practitioners use open-source pre-trained models when they are available.

Despite the ubiquity of using open-source pre-trained models, this practice poses a security threat. Delegating the training process to a third party allows the training party to embed a trigger pattern in the training data. In such a case, the trained model behaves normally in the absence of the trigger but can produce a certain output, determined by the attacker, when the trigger is present. This is known as Trojan or backdoor attacks on machine learning models.

Trojan attacks are hard to detect in a trained model because the model behaves normally on benign inputs. Without knowledge of the trigger, it is impossible to reproduce the model's malicious behavior. Consequently, many proposed methods for Trojan model detection employ reverse engineering to reconstruct possible triggers used to train a given model. The candidate triggers are usually then filtered using heuristics about the trigger size \cite{chenDeepInspectBlackboxTrojan2019}, norm \cite{xiangRevealingBackdoorsPostTraining2020}, or the resulting attack success rate \cite{liuABSScanningNeural2019}. The reverse engineering process can be time-consuming, especially, if it involves attempting all possible combinations of source and target classes for trigger reconstruction \cite{wangNeuralCleanseIdentifying2019}. Furthermore, the deployed heuristics for anomaly detection are susceptible to detecting a trigger when none exists \cite{shenBackdoorScanningDeep2021}.

In this paper, we introduce a novel method for the detection of Trojan models. Our method does not attempt to reconstruct the trigger, nor does it apply heuristics about the nature of the trigger. Instead, we use a purely data-driven approach to detect the presence of a trigger from its fingerprint in the model's signature. The main ingredient of our method is the construction of such a signature for a model, which is accomplished using an activation optimization process that results in a fixed number of activation maps for a given classification model. The signature can be further reduced in size via a feature extraction step that uses pixel-wise statistics. A classifier is then used to detect whether a model is Trojan or not based on the signature or its features. We call our method TRojan Identification from Gradient-based Signatures (TRIGS). The process is illustrated in \cref{fig:framework}. TRIGS is agnostic to the nature of the probe models' architecture. In fact, it works well on very different architectures, as we shall discuss later.

Most of the proposed methods for Trojan model detection in the literature are evaluated on non-public model sets of vastly varying sizes. The few publicly available datasets for image-classification models are limited in the number of classes they support. Also, the vast majority of the model architectures are convolutional. In this paper, we introduce a new dataset of vision transformer (ViT) models~\cite{dosovitskiyImageWorth16x162020} trained on ImageNet. Our dataset will be the largest public dataset in terms of the number of classes (1000) supported by its classification models. It is also the only dataset that focuses on the ViT architecture, which has recently become a popular backbone for many computer vision tasks~\cite{carionEndtoEndObjectDetection2020,xieSegFormerSimpleEfficient2021}. On our collected data and two public datasets, TRIGS delivers state-of-the-art performance.

\begin{figure*}[ht]
    \centering
    \includegraphics[width=\linewidth]{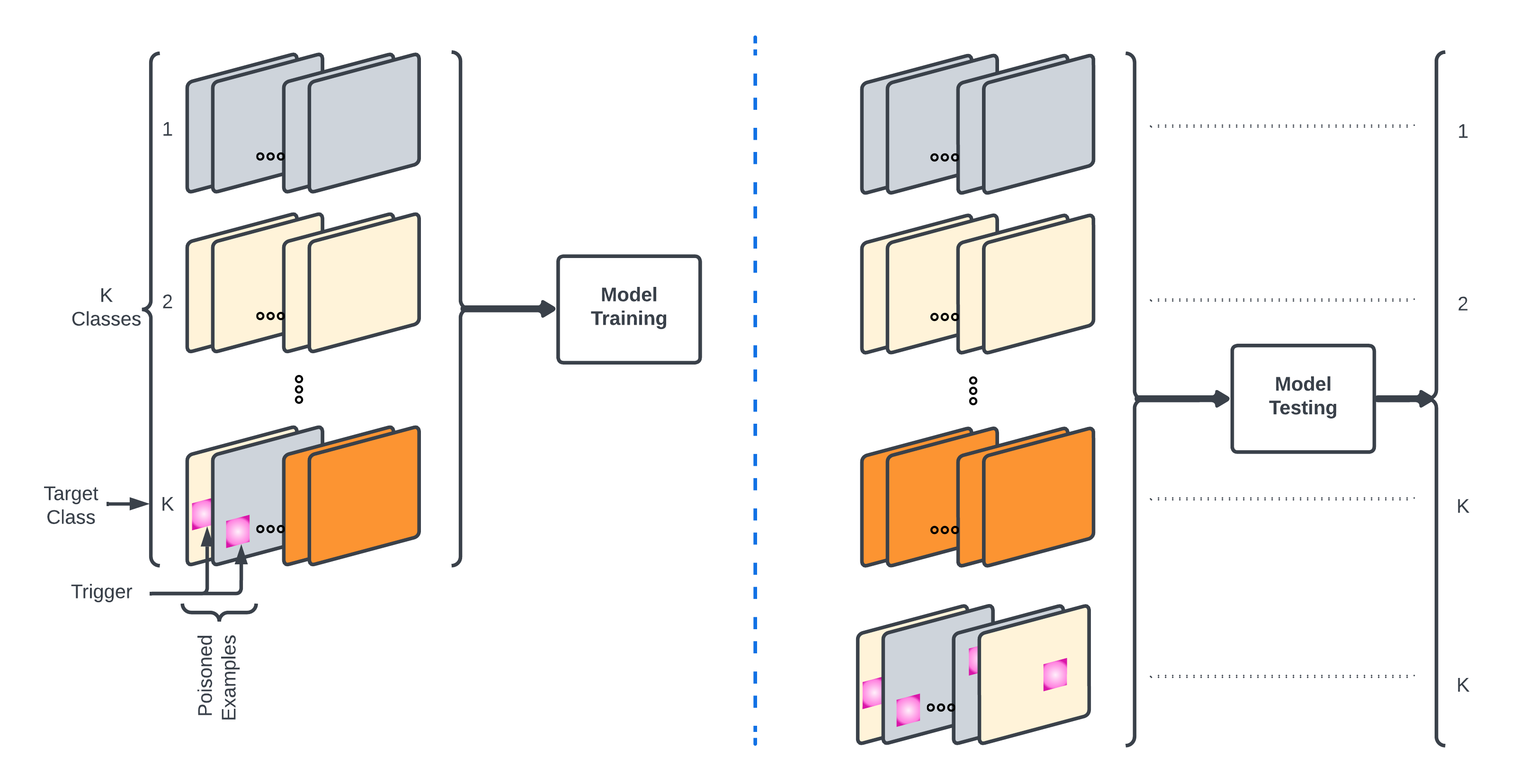}
    \caption{An illustration of universal Trojan attacks. During training, a trigger is embedded in samples from all classes (victim classes) and the contaminated samples are all given one class label (the target class), which is class $K$ in this illustration. During testing, clean inputs are classified correctly. The inputs with the embedded trigger are classified as class $K$.}
    \label{fig:trojan_attack}
\end{figure*}

The contributions of this work can be summarized as follows.

\begin{itemize}
    \item Introducing a novel data-driven method for Trojan model detection based on a fixed-size model signature, regardless of the nature of the probe model's architecture.
    \item Introducing a new dataset for Trojan model detection that is based on the vision transformer architecture and trained on the ImageNet dataset.
    \item Evaluating the performance of our introduced method on our dataset and other public datasets, showing a significant advantage over the baseline methods in both CNN and ViT architectures.
    \item Analyzing and demonstrating the effectiveness of our method even if the defender has access only to much fewer clean examples than the attacker or assumes a different architecture from the one of the attacked model.
\end{itemize}

\section{Related work}
In this section, we discuss the work most related to ours. In \cref{sec:act_max}, we cover the work related to our proposed solution. In \cref{sec:troj_attacks,sec:troj_defense}, we cover Trojan attacks and their defenses in general. For a more comprehensive review of Trojan attacks, the reader is referred to \cite{goldblumDatasetSecurityMachine2023}. Finally, we discuss datasets for Trojan attack detection in \cref{sec:troj_datasets}.

\subsection{Activation maximization}
\label{sec:act_max}
Activation maximization, also known as model inversion or feature visualization, was first introduced in \cite{erhanVisualizingHigherLayerFeatures2009} to visualize the internal nodes of a neural network. The method employed gradient descent with L2 regularization to visualize internal units of Stacked Denoising Auto-encoders and Deep Belief Networks. In \cite{simonyanDeepConvolutionalNetworks2014}, the same technique was applied to convolutional networks. The authors also showed that this gradient-based approach is a generalization of the deconvolution-based approach in~\cite{zeilerVisualizingUnderstandingConvolutional2014}, which was proposed for the same purpose. In~\cite{yosinskiUnderstandingNeuralNetworks2015}, Gaussian blur and pixel clipping were added as additional regularization techniques to produce smoother visualizations. Alternatively to Gaussian blur, in~\cite{mahendranVisualizingDeepConvolutional2016}, random jittering and minimization of the total variation were introduced as extra regularization techniques. More feature visualization techniques are discussed in~\cite{nguyenUnderstandingNeuralNetworks2019a}. It is also worth noting that activation maximization techniques are related to model inversion attacks, in which the attacker's objective is to retrieve private information used in training a model by reconstructing the input that maximizes a certain output, e.g.~\cite{kahlaLabelOnlyModelInversion2022,wangVariationalModelInversion2022}.

\subsection{Trojan attacks}
\label{sec:troj_attacks}
Trojan attacks on deep learning models were first introduced by \cite{guBadNetsIdentifyingVulnerabilities2019}, in which mislabeled examples stamped with a trigger were used to train a Trojan model. In \cite{liuTrojaningAttackNeural2018}, a method was presented for creating Trojan attacks without access to training data by using model inversion~\cite{nguyenMultifacetedFeatureVisualization2016}. In~\cite{shafahiPoisonFrogsTargeted2018,geipingWitchesBrewIndustrial2022}, methods for clean label poisoning attacks were introduced. These methods target the misclassification of a specific test example.
Interestingly, in~\cite{sahaHiddenTriggerBackdoor2020,souriSleeperAgentScalable2022a}, clean label attacks were carried out in such a way that a trigger can be used in the testing phase while being completely hidden during training.
Dynamically generated triggers, which are based on the input sample, rather than being fixed, were introduced in~\cite{nguyenInputAwareDynamicBackdoor2020}. In ~\cite{ditangDemonVariantStatistical2021}, a method for source-specific attacks was introduced and was shown to be more resistant to detection. The method works by adding cover images, which are images that include the trigger but are given the correct label.

\subsection{Defenses against Trojan attacks}
\label{sec:troj_defense}

Defenses against Trojan attacks include the detection of poisoned samples in a training dataset \cite{tranSpectralSignaturesBackdoor2018,chenDetectingBackdoorAttacks2019} and making model training robust against poisoned samples \cite{weberRABProvableRobustness2022,levineDeepPartitionAggregation2021}. In both types of defenses, it is assumed that the defender has control over the training process. Other types of defenses include modifying a known Trojan model to bypass the trigger \cite{wuAdversarialNeuronPruning2021,liNeuralAttentionDistillation2022} and detecting if a trained model is Trojan or not \cite{chenDeepInspectBlackboxTrojan2019,kolouriUniversalLitmusPatterns2020}. Our focus in this section is on the latter type of defense, which is the topic of this paper. We argue that Trojan model detection is an indispensable capability because it is the first step towards removing the effect of the trigger if it is present. If the detection is incorrect, either the probe model's performance will be unnecessarily compromised as a side effect of attempting to remove a non-existing trigger, in the case of a false positive; or the probe model will be used while being infected, in the case of a false negative. Therefore, Trojan model detection research is very relevant and important.

DeepInspect~\cite{chenDeepInspectBlackboxTrojan2019} is an algorithm for detecting models with backdoors assuming that the defender has access only to the trained model and no access to clean data samples. To achieve this goal, the method uses model inversion~\cite{fredriksonModelInversionAttacks2015} to construct a training dataset for the model. Using the constructed training data, a generator is employed to create perturbation patterns (triggers) such that the model produces a given target class with the poisoned samples. Then, anomaly detection is applied to determine if any of the generated triggers is a real trigger used to train the model. Similarly, in \cite{wangNeuralCleanseIdentifying2019}\cite{xiangRevealingBackdoorsPostTraining2020}, anomaly detection is applied to detect the real trigger (if any) among a set of generated triggers. The idea was extended in~\cite{dongBlackboxDetectionBackdoor2021} to the black-box case, where the model is only accessible through its query responses. However, in this case, the triggers are generated by reverse engineering with real clean samples. In~\cite{liuABSScanningNeural2019}\cite{shenBackdoorScanningDeep2021}, potentially compromised neurons are first identified. Based on the identified compromised neurons, possible triggers are generated and only those that consistently subvert the model's predictions to a certain target class are admitted. The methods require at least one clean sample of each class. The case when the clean samples available to the defender are limited or non-existent was handled in~\cite{wangPracticalDetectionTrojan2020}. The method uses the similarity between two embeddings per image, one with a universal perturbation pattern and one with a local perturbation pattern, as an indicator of the presence of a Trojan. Universal Litmus Patterns (ULPs)~\cite{kolouriUniversalLitmusPatterns2020} are input patterns to a classification model, the output of which can be used to distinguish benign from Trojan models. Few patterns per class have been shown to be effective in detecting Trojan models on multiple datasets. A very similar idea was proposed and thoroughly analyzed in~\cite{wangDetectingAITrojans2021}. Complex attack scenarios, in which the trigger pattern is not limited to be patch-shaped, were the focus of~\cite{liuComplexBackdoorDetection2022}. More recently~\cite{wangMMBDPostTrainingDetection2023}, a detection method was introduced based on the observation that Trojan models have an anomalously large logit margin for the target class. Our proposed method, TRIGs, works both in the white-box and black-box settings and with a limited access to clean data. Unlike recent defenses, which are customized for a specific architecture~\cite{doanDefendingBackdoorAttacks2023}, TRIGS is generic and works well with both CNN and ViT architectures. The closest defense to TRIGS is the One-Pixel Signature (OPS) defense~\cite{huangOnePixelSignatureCharacterizing2020}, in which a model signature is used to train a binary classifier to distinguish Trojan from benign models. However, to work in the black-box scenario, OPS uses brute force search to construct the signatures instead of using gradient descent optimization as in TRIGS. Also, the signature size in OPS is proportional to the number of classes, which can be very large, while TRIGS can leverage pixel statistics to significantly reduce the signature size regardless of the number of classes.

\subsection{Datasets for Trojan attack defense}
\label{sec:troj_datasets}
Unfortunately, most of the work done on Trojan attack defenses used private datasets, usually containing a small number of models. To our knowledge, the only work with models publicly released is the universal litmus patterns work~\cite{kolouriUniversalLitmusPatterns2020}, where the models for the CIFAR10 and Tiny ImageNet classification tasks have been released. More recently, under IARPA's TrojAI program\footnote{\url{https://www.iarpa.gov/research-programs/trojai}}, a software package~\cite{karraTrojAISoftwareFramework2020} and multiple datasets have been released for different computer vision and NLP tasks. Our focus in this paper is on the image classification task in natural images, as opposed to synthetic images used in the TrojAI data collections. The datasets released so far for image classification have been limited in the number of classes supported (maximum is 200 classes in the Tiny ImageNet classification task). Furthermore, there has been no sufficient focus on the vision transformer architecture~\cite{dosovitskiyImageWorth16x162020} despite its rising popularity. Therefore, we create a new dataset based on vision transformer models trained on the ImageNet dataset (1000 classes).

\section{Approach}
\subsection{Threat model}
The attacker is assumed to train a $K$-class classifier and provide it to the victim such that the classifier works normally on clean inputs, but once a trigger is attached to an input, the classifier produces a certain class (the target class) of the attacker's choice. The trigger is assumed to be small in size with respect to the input so that the attacker can deploy the attack in the physical world. The attacker achieves their goal by poisoning a fraction of the training dataset, which is done by adding the trigger to the poisoned fraction from all classes and giving them the target label as the ground truth label during training.
This process is illustrated in \cref{fig:trojan_attack}.
Alternatively, the attacker can release a poisoned dataset to the public such that the victim can train the classifier on their end. In this case, the attacker can choose to use a clean-label poisoning mechanism that still allows the attacker to deploy the attack in the physical world.

The defender, who can be a third party different from the victim, has access to the trained model's weights and hence can use gradient descent to create a signature for the model without the need for any data samples. The defender also can train a binary classifier (\textit{a detector}) that can tell from the signature whether the model is Trojan or not. The detector is trained on signatures from a set of benign and Trojan models for the target $K$-class classification task. To train the detector, the defender needs access to pre-trained benign and Trojan models, which can be obtained from trusted sources, such as NIST's TrojAI data, or can be created by the defender by training a small number of \textit{shadow models} on a small set of clean data.

A similar threat model in the black box setting was used in~\cite{kolouriUniversalLitmusPatterns2020,huangOnePixelSignatureCharacterizing2020,wangDetectingAITrojans2021}. We show that our approach still works in the black-box setting. However, it is important to note that targeting the white box case is still practical due to the wide-spread use of pretrained model weights downloaded from the web. In such cases, when the model weights are available, it is imperative to leverage them to enhance the detectability of Trojan models.

\subsection{Intuition}
Due to the way the attack is installed, the Trojan model develops a strong association between the trigger pattern and the target class. Such a strong association is expected to be evident upon model inversion. Namely, if we attempt to synthesize an image that maximizes or minimizes the activation associated with the target class, the trigger pattern is expected to have a fingerprint in such an image. Not only that, but the trigger's fingerprint is expected to appear even if we are maximizing or minimizing the activations of other classes. For example, if our objective is to minimize the activation of a class other than the target one, the easiest way could be just to add the trigger to an image. Similarly, if the objective is to maximize such an activation instead, the model would make sure that it does not have any trace of the trigger. Therefore, whether we are maximizing or minimizing the activation of any class, the trigger can have a fingerprint on the resulting image.

\subsection{Framework}
\Cref{fig:framework} illustrates the proposed framework, which generalizes the intuition outlined above. Given a trained $K$-class classifier, a signature is created by finding images that optimize $M$ loss functions, which are computed based on the logits of the $K$ classes. Therefore, $M$ is a function of $K$. This results in $M$ such images, which collectively constitute the signature for the model. A classifier is then trained to determine from the model's signature whether it is Trojan or not, after an optional feature extraction step. 

\subsection{Activation optimization}
\label{sec:act_opt}
Let $f(x)$ be a $K$-class classification model. That is, $f: \mathbf{R}^{C\times H \times W} \rightarrow \mathbf{R}^K$, such that the input to the function $f$ is a $C$-channel $H\times W$ image, and the output is a vector of $K$ logits corresponding to the $K$ classes. The $i^{th}$ activation optimization map of the signature is defined as
\begin{equation}
    a_i = \argmin_x L_i(f(x)) \enspace ,
\end{equation}
where $L_i$ is a loss function defined over the logits corresponding to an input $x$.
Then the signature of the model is defined as
\begin{equation}
    \mathcal{S} = \left[a_1 | a_2 | \dots | a_{M-1} | a_M \right] \enspace ,
\label{eq:signature}
\end{equation}
where $|$ is the channel-wise image concatenation operator.

In the current realization of our framework, we use $M\leq 2K$ loss functions, where $M=K$ when we use logit minimization or maximization as our loss functions, and $M=2K$ when we combine logit maximization and minimization together. Let $f_j(x)$ be the $j^{th}$ element of the output of $f$. In the case of combining minimization and maximization, the $i^{th}$ loss function is defined as
\begin{equation}
    L_i(f(x))= 
    \begin{cases}
        f_i(x)      & i \in \mathbf{Z^+}, i \leq K \\
        -f_{i-K}(x) & i \in \mathbf{Z^+}, K <i  \leq 2K
    \end{cases} \enspace .
\end{equation}
For the rest of the paper, unless otherwise specified, we will use the variant of the signature with $M=2K$.

\begin{figure*}
    \centering
    \includegraphics[width=\linewidth]{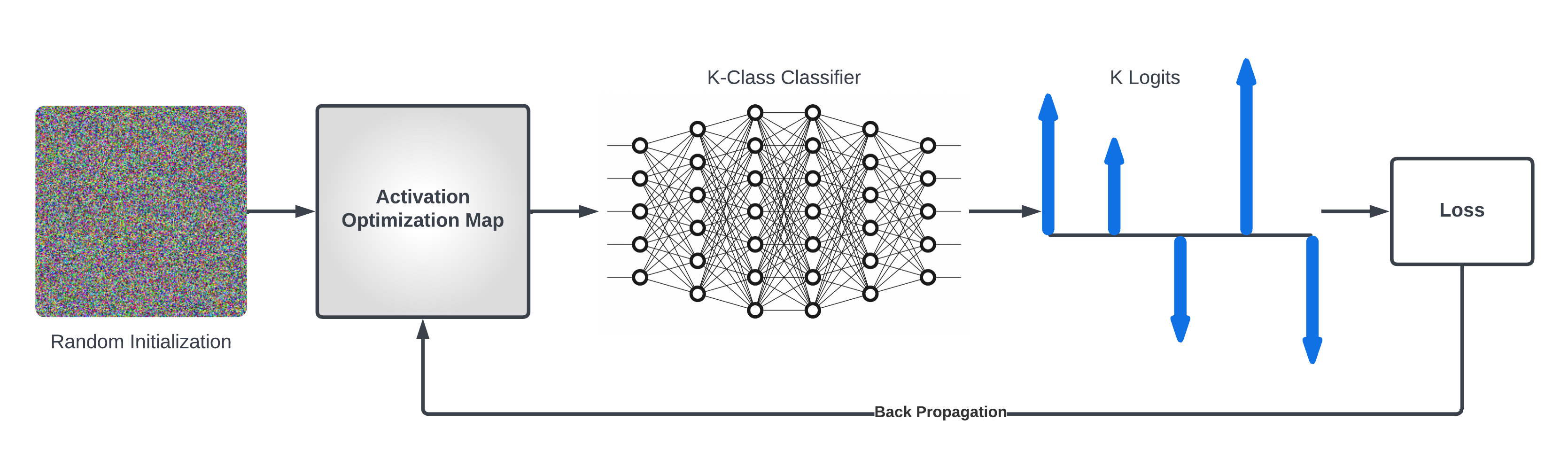}
    \caption{The activation optimization process. Starting from a random image, an activation optimization map is derived using gradient descent on a loss function based on the classification logits.}
    \label{fig:activation_opt}
\end{figure*}

\subsubsection{Regularization}
The activation optimization process can be implemented using gradient descent starting from a random image, as shown in \cref{fig:activation_opt}. However, a number of regularizations are important to make the resulting images as natural as possible. Otherwise, we may end up having images that contain no useful patterns. In particular, we applied the following regularization techniques during activation optimization.

\paragraph{$L_2$ regularization} This is the most common regularization technique used in model inversion. It works by adding the $L_2$ norm of the resulting image as a term in the loss. That is
\begin{equation}
    R_{L_2}(x) = || x ||_2 \enspace .
\end{equation}

\paragraph{Total variation regularization} The total variation regularization~\cite{mahendranVisualizingDeepConvolutional2016} is used to enhance the smoothness of the generated image by minimizing the local gradients at every pixel. In particular, we minimize the $L_1$ norm of the local gradient in each channel as follows.
\begin{multline}
    R_{TV}(x) = \sum_{ijk} |x(i, j, k) - x(i, j-1, k)| \\ + |x(i,j, k) - x(i-1, j, k) | \enspace ,    
\end{multline}
where $x(i, j, k)$ is the pixel value at location $(i,j)$ in the $k^{th}$ channel of $x$.

Adding the main loss and the regularization terms together, the $i^{th}$ activation optimization map is obtained by
\begin{equation}
    a_i = \argmin_x L_i(f(x)) + \lambda_{L_2}R_{L_2}(x) + \lambda_{TV} R_{TV}(x) \enspace ,
\end{equation}
where $\lambda_{L_2}$ and $\lambda_{TV}$ are loss term weight parameters to be finetuned.

    
    

\subsection{Feature extraction}
\label{sec:feat_ext}
The size of our constructed model signature grows linearly with the number of classes. When the number of classes is large, training a classifier on the resulting signature may not be practical. To address this issue, we propose a feature extraction step, which converts the signature into a fixed number of channels regardless of the number of classes. The idea is to use pixel-wise statistics over the signature channels. Consider the signature $\mathcal{S}$ composed of $M$ activation optimization maps, as shown in \cref{eq:signature}. Suppose that each activation optimization map contains $c$ channels (typically $c=3$). Then, $\mathcal{S}$ has $N=cM$ channels in total. Consider the pixel at $(i,j)$ in the $N$ channels of $\mathcal{S}$. Let $\text{s}_{ij}=[s_{ij1}s_{ij2}\dots s_{ijN}]$ be a vector containing the values of the $N$ channels at the $(i,j)$ pixel location. Let $g: \mathbf{R}^N \rightarrow \mathbf{R}^P$ such that $\text{u}_{ij}=g\left(\text{s}_{ij}\right)$ be a vector of $P$ statistics computed over the values of $\text{s}_{ij}$. The compilation of the pixel statistics vectors constitute a $P$-channel feature map $\mathcal{U}$ whose size is independent of the number of maps $M$ in the raw signature $\mathcal{S}$. Specifically, we set $P=11$, where the 11 statistics are as follows: minimum, maximum, sample mean, sample standard deviation, 0.25 quantile, median, 0.75 quantile, and four histogram bins.

As discussed in \cref{sec:act_opt}, our current realization uses a combination of activation minimization and activation maximization maps. That is $\mathcal{S}=\left[\mathcal{S}_{\text{min}}| \mathcal{S}_{\text{max}}\right]$, where $\mathcal{S}_{\text{min}}$ and $\mathcal{S}_{\text{max}}$ are the portions of $\mathcal{S}$ that correspond to the activation minimization maps and the activation maximization maps, respectively. In addition to the pixel statistics feature map $\mathcal{U}$, whose values are computed over all channels of $\mathcal{S}$, we also construct $\mathcal{U}_\text{min}$ and $\mathcal{U}_\text{max}$, which are pixel statistics feature maps computed over the channels of $\mathcal{S}_\text{min}$ and $\mathcal{S}_\text{max}$, respectively. Therefore, our final pixel statistics feature map is $\left[\mathcal{U}_\text{min}|\mathcal{U}_\text{max}|\mathcal{U}\right]$ with a total of 33 channels.

\subsection{Detection}
\label{sec:detection}
For deciding if a model's signature or its derived feature map corresponds to a Trojan model or not, we need to train a binary classifier. For this classifier, we employ a convolutional neural network architecture. Specifically, we use a ResNeXt-50 (32x4d) \cite{xie2017aggregatedResNeXt} architecture with the first layer of the model modified to accept the number of channels in the input.

\section{Experimental evaluation}
\subsection{Evaluation data}
\paragraph{Public datasets} To evaluate our method, we use two public datasets introduced in~\cite{kolouriUniversalLitmusPatterns2020} and create our own dataset, which we will make publicly available. One of the two public datasets is for models trained on the CIFAR10 dataset. The models are based on a modified version of the VGG architecture \cite{Simonyan15VGG}. The other public dataset is for models trained on the Tiny ImageNet dataset. The models for the latter dataset are based on a shallow version of the ResNet18 architecture \cite{he16Resnet}, which we will refer to as ResNet10. In both datasets, 20 different trigger patterns were used such that 10 of them appear only in the training models, and the other 10 appear only in the testing models. The numbers of samples in each split of the two datasets are shown in \cref{tab:datasets}. More details about the datasets can be found in~\cite{kolouriUniversalLitmusPatterns2020}.

\begin{figure}
    \centering
    \includegraphics[width=\linewidth]{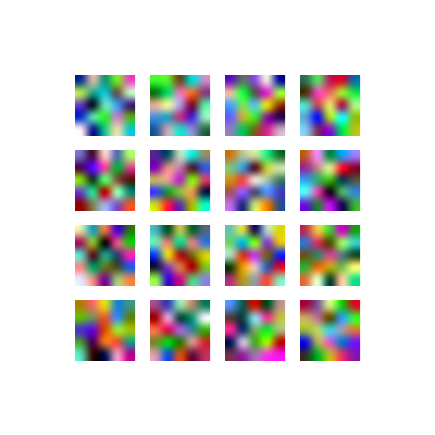}
    \caption{Sample triggers from our dataset. Each trigger is $32\times 32$. Triggers are created by resizing $5\times 5$ random patches to $32\times 32$ using bicubic interpolation. Each Trojan model is trained with a unique trigger.}
    \label{fig:troj_triggers}
\end{figure}

\paragraph{Our dataset} Our own collected dataset contains $1,200$ models, with $600$ benign and $600$ Trojan. From each class, we use 500 models for training and the remaining 100 for testing, as depicted in \cref{tab:datasets}. All models in our dataset are created from a pre-trained ViT-B-16 architecture~\cite{dosovitskiyImageWorth16x162020} available with the \texttt{torchvision} package~\cite{paszkePyTorchImperativeStyle2019}. Specifically, we used the weight version named \texttt{ViT\_B\_16\_Weights.IMAGENET1K\_V1}. Each model was then trained for one epoch on 90\% of the ImageNet training set using the AdamW optimizer~\cite{loshchilovDecoupledWeightDecay2018} with a learning rate of $10^{-5}$ and a batch size of 64.
For each Trojan model, a random target class was chosen, and a randomly generated trigger was created and placed at a random location in 1\% of the training data. A trigger was generated by first randomly sampling a $5\times 5$ 3-channel tensor and then resizing it to $32\times 32$ using bicubic interpolation.
Example generated triggers are shown in \cref{fig:troj_triggers}.
The performance of the original ViT-B-16 model on the ImageNet validation data was 81\%. After training for one epoch, the accuracy of our benign models dropped to around 79\% (which could be due to overfitting), and the accuracy of the poisoned models on clean data was between 78\% and 79\%. Therefore, our Trojan models preserved performance on clean data. On the other hand, with the addition of triggers, the performance of Trojan models dropped to almost 0\% in all victim classes, which means that the trigger was effective in poisoning the model.

\begin{table}[]
\caption{Construction of our evaluation datasets. The CIFAR10 and Tiny ImageNet datasets are obtained from~\cite{kolouriUniversalLitmusPatterns2020}. The ImageNet dataset is created by us and will be publicly released.}
\label{tab:datasets}
\centering
\begin{tabular}{lcc>{\centering}p{1.3 cm}>{\centering\arraybackslash}p{1.3 cm}}
\toprule
\textbf{Dataset}   & \textbf{Arch}      & \textbf{Split}               & \textbf{Benign Models}     & \textbf{Trojan Models}  \\ \midrule
CIFAR10  & VGG        & Train & 500  & 500  \\
CIFAR10  & VGG        & Test  & 100  & 100  \\
Tiny ImageNet & ResNet  & Train & 1000 & 1000 \\
Tiny ImageNet & ResNet   & Test  & 100  & 100 \\
ImageNet & ViT       & Train & 500  & 500 \\
ImageNet & ViT       & Test  & 100  & 100 \\ \bottomrule
\end{tabular}
\end{table}

\begin{figure}
    \centering
    \includegraphics[width=\linewidth]{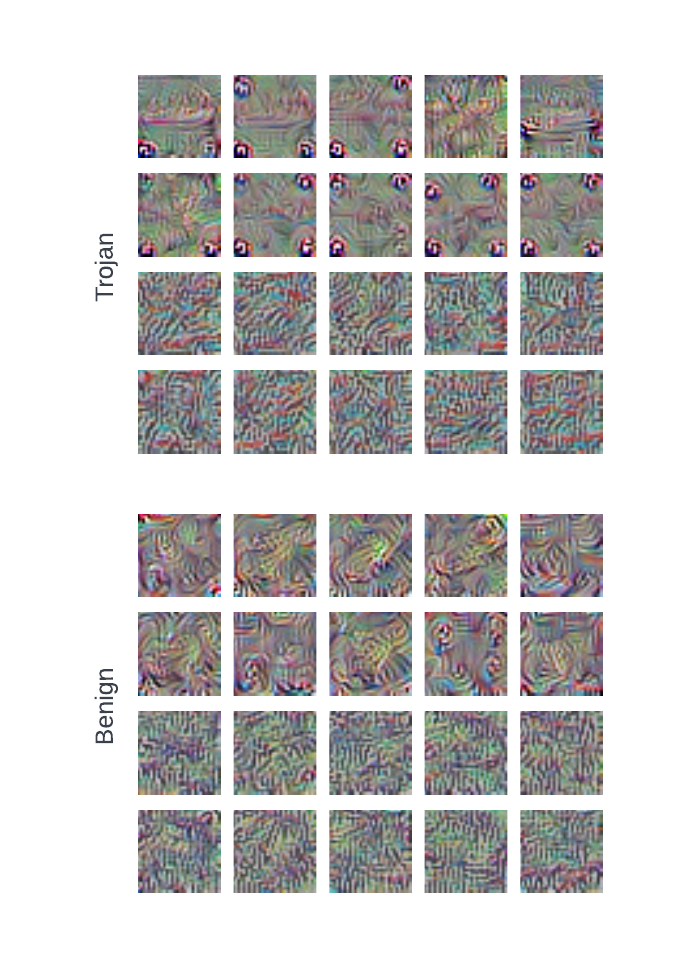}
    \caption{Sample signatures from the CIFAR10 Trojan dataset. Each signature has 20 images corresponding to the 10 classes of the dataset. The top two rows of each signature are for the activation minimization maps while the bottom two rows are for the activation maximization maps. Note how the trigger has a clear fingerprint in the minimization maps for the signature of the Trojan model.}
    \label{fig:signature_sample}
\end{figure}

\subsection{Implementation details}
\label{sec:impl}
Signatures were created using the Adam optimizer~\cite{kingmaAdamMethodStochastic2015} with 200 iterations. A learning rate of 10 was used with the CIFAR10 dataset while a learning rate of 0.1 was used with the Tiny ImageNet and the ImageNet datasets. For the CIFAR10 dataset, it was important to standardize the final image so that it has pixel values with 0.5 mean and 0.25 standard deviation.

$L_2$ regularization was implemented by setting the weight decay argument of the optimizer to $10^{-5}$. The weight for the total variaton regularization was set to $10^{-3}$ for the CIFAR10 and ImageNet datasets, and was set to $10^{-2}$ for the Tiny ImageNet dataset.

The detection classifier model was trained using the Adam optimizer with a learning rate of $10^{-4}$, and with 100 epochs. 90\% of the training samples were used for training and the remaining 10\% were used for validation.

\subsection{Evaluation results}
\label{sec:results}

Sample signatures for one Trojan model and one benign model from the CIFAR10 Trojan dataset are shown in \cref{fig:signature_sample}. In this dataset, the trigger was placed at the corners of the image. You can see the footprint of the trigger clearly at the corners of the Trojan model's signature, particularly in the activation minimization maps (top two rows).


In the remainder of this section, we focus on comparing different variants of our methods to prior research.
\Cref{tab:perf} shows the area under the receiver operating characteristics curves (AUC) for detecting Trojan attacks using our method and two baseline methods on the three datasets. 
As explained in \cref{sec:act_opt}, we applied three variants of our method using activation minimization, activation maximization, and both, in which case we concatenated the signature channels coming from the former two optimizations. We also used the pixel statistics channels as explained in \cref{sec:feat_ext}. In \cref{tab:perf}, for all experiments on CIFAR10 and Tiny ImageNet and for the statistics experiment on ImageNet, we present the average and the standard deviation of the AUC score over ten independent training sessions for each classifier. For activation maximization, minimization, and their combination on ImageNet, we only trained three models for each due to the heavy computational cost.

The three baseline methods, which are ULP~\cite{kolouriUniversalLitmusPatterns2020}, k-Arm optimization~\cite{shenBackdoorScanningDeep2021}, and MM-BD~\cite{wangMMBDPostTrainingDetection2023}, were chosen based on the availability of their code and its adaptability to new datasets. For ULP, we used the publicly available code for CIFAR10 and Tiny ImageNet. We applied the best configuration in the paper for each dataset, which was ten litmus patterns for CIFAR10 and five for Tiny ImageNet. We created ten sets of litmus patterns for each dataset. In \cref{tab:perf}, we report the mean and standard deviation of the AUC and accuracy scores over the litmus pattern sets. It is worth noting that we could not reproduce or come close to the results reported in the ULP paper despite using the code released by the authors. For the ImageNet dataset, we could not get the method to work due to excessive computational cost and lack of convergence.

\begin{figure}
    \centering
    \includegraphics[width=1\linewidth]{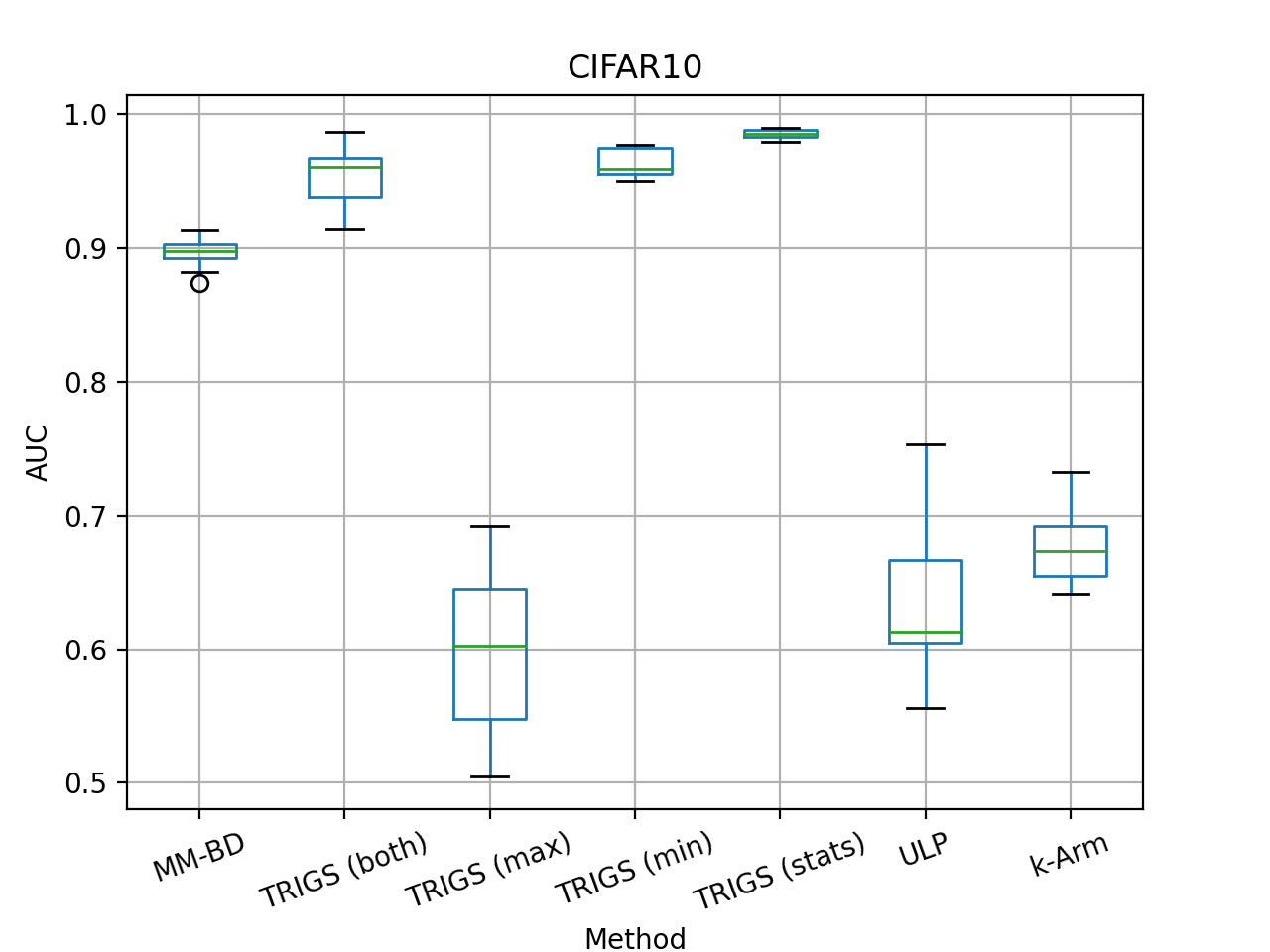}
    \caption{AUC whisker plots for CIFAR10}
    \label{fig:cifar10-wp}
\end{figure}

\begin{figure}
    \centering
    \includegraphics[width=1\linewidth]{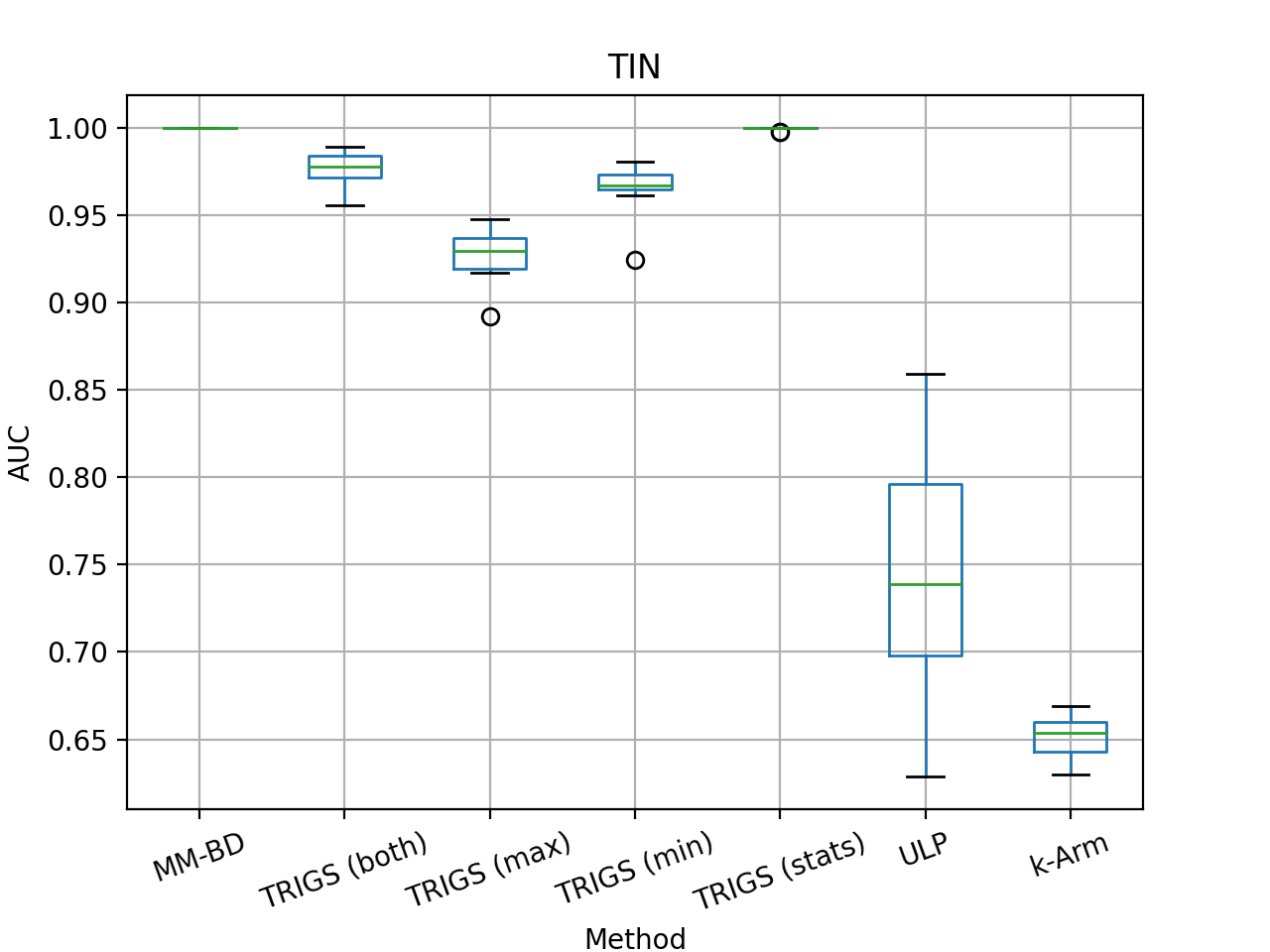}
    \caption{AUC whisker plots for Tiny ImageNet}
    \label{fig:tin-wp}
\end{figure}

\begin{figure}
    \centering
    \includegraphics[width=1\linewidth]{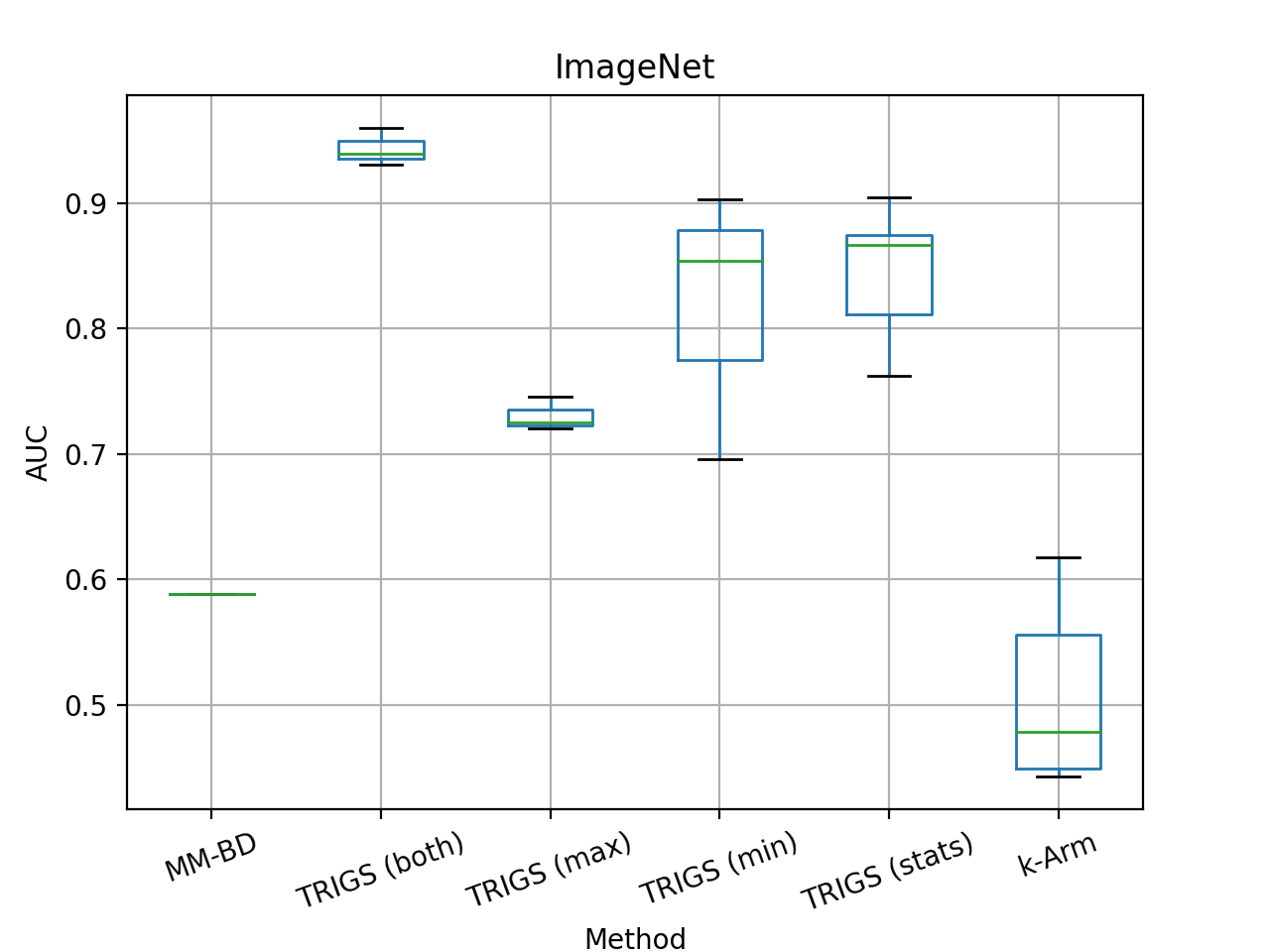}
    \caption{AUC whisker plots for ImageNet}
    \label{fig:imagenet-wp}
\end{figure}

For k-Arm optimization, we adapted the publicly released implementation and evaluated it on the three datasets. We used the Trigger Size output for each model as the score based on which we computed the AUC. Again, we evaluated the method ten times for each probe model with different random seeds. We report the mean and standard deviation of the resulting AUC scores in \cref{tab:perf}. In \cref{fig:cifar10-wp,fig:tin-wp,fig:imagenet-wp}, box plots are used to present the AUC scores for all the runs.

For the MM-BD method, we adapted the publicly available code to work with our datasets. We found that the default number of steps used in the paper (300) was too small for the models to converge. For a fair comparison, in all our experiments, we let the optimization run until convergence. We used ten different runs for each model in the CIFAR10 and the Tiny ImageNet datasets. However, due to the excessive computational time, we only used one run for the ImageNet dataset.

From the results in \cref{tab:perf} and \cref{fig:cifar10-wp,fig:tin-wp,fig:imagenet-wp}, we can observe that, in each dataset, at least one of our four variants surpasses or matches the baseline performance, regardless of whether the probe model is CNN or ViT-based. Moreover, when our method surpasses the baseline methods, the margin is statistically significant.

It is interesting to observe that the pixel-wise statistics variant is the only variant that consistently outperforms or matches all baseline methods. It is also the best in the case of CIFAR10 and Tiny ImageNet, achieving the highest mean score and the lowest standard deviation. However, for ImageNet, the variant that combines both types of activation optimization maps achieves the best performance. It is also interesting to notice that the activation minimization variant consistently performs better than the activation maximization one. This result is surprising given that all prior work on model inversion focused on activation maximization (or alternatively minimizing the classification loss, e.g. the cross-entropy loss). Here, for the first time, we find a good use for activation minimization-based model inversion.

Out of the three baseline methods, the only serious contender is the MM-BD method. In fact, this method achieves a perfect AUC score on the Tiny ImageNet dataset (though its accuracy is not the best). However, similar to the other two baseline methods, MM-BD struggles on the ImageNet dataset. We believe this struggle is due to the ViT architecture, in which the main assumption of the MM-BD method (the presence of an anomalously large logit margin for the target class in a Trojan model) may not hold.

\begin{table*}[]
\small
\caption{Performance results as areas under the ROC curves (AUC) for baseline models and the four variants of our method. When numbers are included between parentheses, they represent the standard deviation over multiple runs (10 for most of the cases, as explained in the text) while the value outside the parentheses are the averages over the same runs.}
\label{tab:perf}
\centering
\begin{tabular}{lccccccc}
\toprule
      &                      & \multicolumn{2}{c}{CIFAR10}                              & \multicolumn{2}{c}{Tiny ImageNet}                                 & \multicolumn{2}{c}{ImageNet}                     \\
      &  & AUC          & Acc                   & AUC                   & Acc                   & AUC  & Acc                   \\ \midrule
ULP   &                      & 0.64 (0.060)                     & 0.61 (0.048)          & 0.74 (0.075)                              & 0.71 (0.066)          & --                       & --                    \\
k-Arm &                      & 0.68 (0.028)                     & 0.51 (0.000)          & 0.65 (0.120)                              & 0.54 (0.024)          & 0.51 (0.67)              & 0.5 (0.000)           \\
MM-BD &  & 0.90 (0.012) & 0.79 (0.029)          & \textbf{1.00 (0.000)} & 0.97 (0.009)          & 0.59 & 0.51                  \\ \midrule
\multirow{4}{*}{TRIGS} & Both                 & 0.95 (0.022)                     & 0.90 (0.038)          & 0.98 (0.010)                              & 0.93 (0.014)          & \textbf{0.94 (0.015)}    & \textbf{0.87 (0.033)} \\
      & Max                  & 0.60 (0.067)                     & 0.57 (0.054)          & 0.93 (0.016)                              & 0.83 (0.047)          & 0.73 (0.013)             & 0.66 (0.020)          \\
 & Min                  & 0.96 (0.011)                     & 0.92 (0.019)          & 0.96 (0.015)                              & 0.92 (0.013)          & 0.82 (0.108)             & 0.75 (0.083)          \\
      & Stats                & \textbf{0.99 (0.003)}            & \textbf{0.96 (0.011)} & \textbf{1.00 (0.001)}                     & \textbf{0.99 (0.010)} & 0.84 (0.046)             & 0.76 (0.050)          \\ \bottomrule
\end{tabular}
\end{table*}

\subsection{Sensitivity to chosen statistics}
\begin{figure}
    \centering
    \includegraphics[width=1\linewidth]{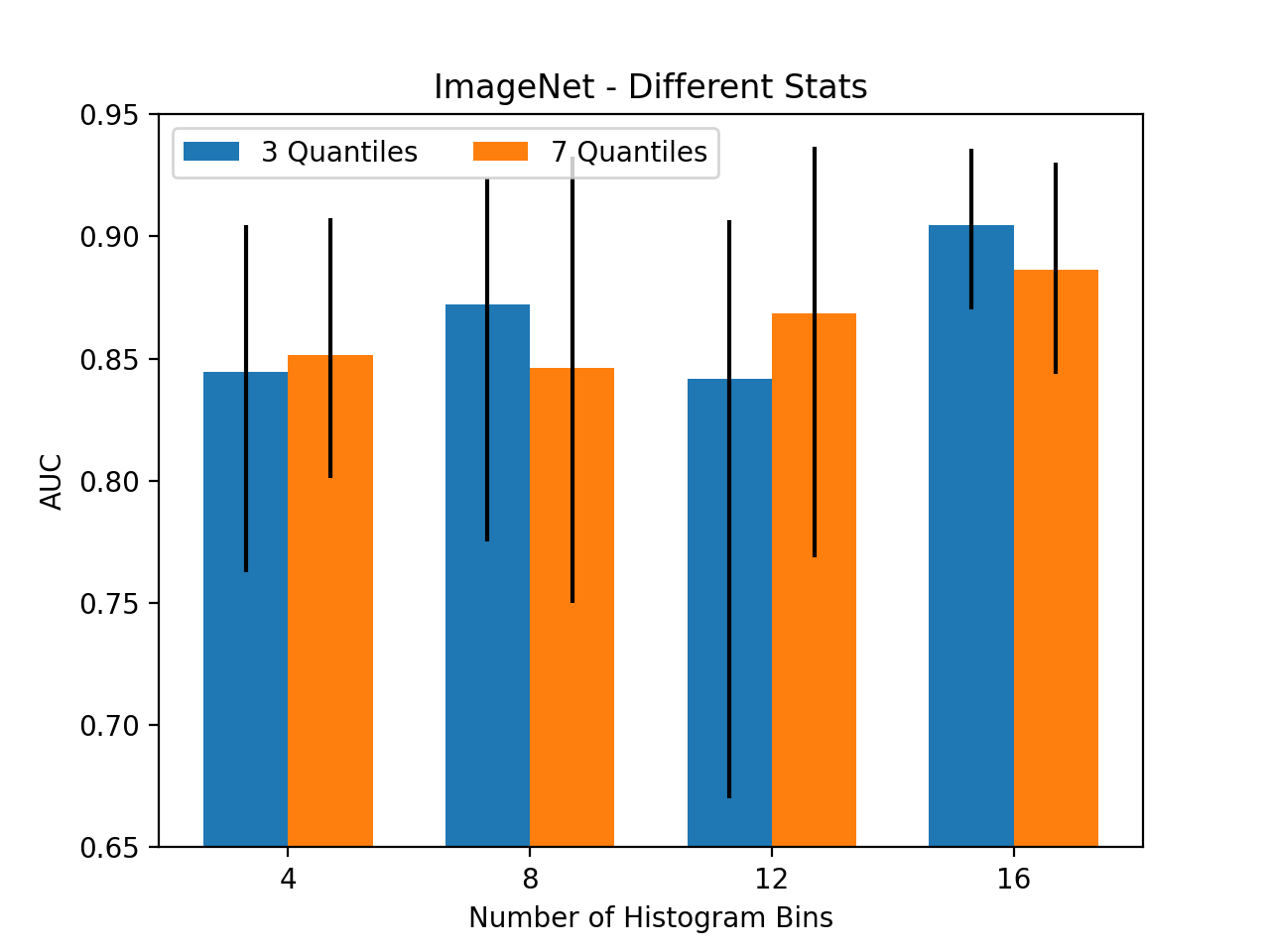}
    \caption{Average AUC with different number of histogram bins and quantiles for the ImageNet dataset. The error lines show the range of values.}
    \label{fig:imagenet-ablation}
\end{figure}

Since the pixel-wise statistics variant is the most efficient and provides the best performance (if we consider all datasets together and accounting for the computational efficiency), we study in this section the sensitivity of the method to varying the number of used statistics. We focus on the ImageNet dataset here because the other two datasets are almost saturated. The results in \cref{tab:perf} are for 11 statistics, as explained in \cref{sec:feat_ext}. We experimented with adding more quantiles. Specifically, instead of using three quantiles at 0.25, 0.5 (median), and 0.75; we added four more at 0.125, 0.375, 0.625, 0.875. We also experimented with different numbers of histogram bins. \Cref{fig:imagenet-ablation} shows the results of these experiments. The bars represent the mean AUC over 10 training session and the error lines represent the range of values. There is no clear advantage of adding more quantiles. However, using more histogram bins can slightly enhance the performance at the cost of more memory and computational cost.

\subsection{Stronger threat models}
In this section, we study the effect of having a stronger threat model. In particular, we study three aspects of the threat model: (1) the data available to the defender for training shadow models is different and much smaller than the data available to the attacker, (2) the defender uses a different architecture to train the shadow models, and (3) the defender uses a small number of shadow models.

To conduct these experiments, we created another set of models trained on the Tiny ImageNet dataset. To mimic the effect of having different and smaller data available to the defender, we split the dataset into two disjoint sets: a large set consisting of 50\% of the original training data used only by the attacker (i.e. the testing models.) and 4-10\% of the data used only by the defender to train the shadow models. All shadow models were trained on the ResNet10 architecture adopted in~\cite{kolouriUniversalLitmusPatterns2020}. For each percentage of the data used to train the shadow models, we trained 1000 of them, split as 500 benign and 500 Trojan models. For the testing models (representing the attacker's trained models), we trained 200 models using the ResNet10 architecture and 200 models using the VGG16 architecture. For each architecture, half of the models were benign and the other half were Trojan. Each model, whether used for training or testing the detector, was trained on a unique random trigger created in a similar way to what we used for the ImageNet-ViT dataset, but using a trigger size of $8\times 8$. Triggers are placed in random locations in 2\% of the training data in the case of the testing models, and in 5\% of the training data in the case of the training models. The reason for having different poisoning fractions is that as the size of the training data reduces, we found that a higher poisoning fraction is needed to achieve a high attack success rate (typically $\sim 98\%$).

\begin{figure}
    \centering
    \includegraphics[width=1\linewidth]{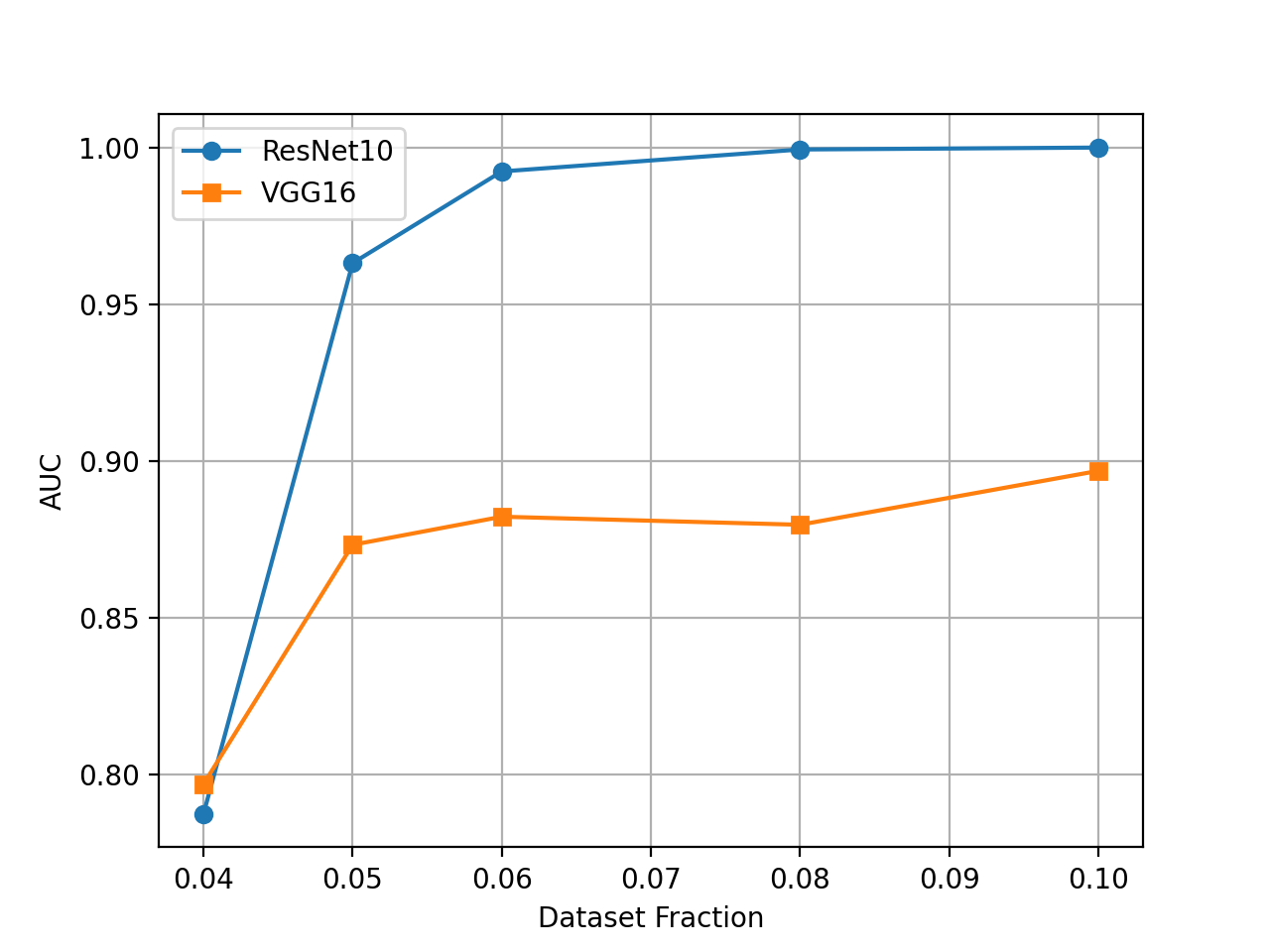}
    \caption{Average AUC with different fractions of the Tiny ImageNet dataset used to train the shadow models. All the testing models were trained on 50\% of the data using two different architectures, ResNet10 and VGG16.}
    \label{fig:dataset_frac}
\end{figure}

\Cref{fig:dataset_frac} shows the average AUC plots for these experiments. Each point is an average of 10 different runs. For these experiments, we used the pixel-wise statistics variant of our method. As can be observed from the plots, as low as 6\% of the dataset is enough for excellent performance if the architecture of the shadow models matches with that of the probe models. When the architectures are different, despite the drop in performance, it is still higher than the baseline methods, ranging from around 0.8 to 0.9 AUC.

\begin{figure}
    \centering
    \includegraphics[width=1\linewidth]{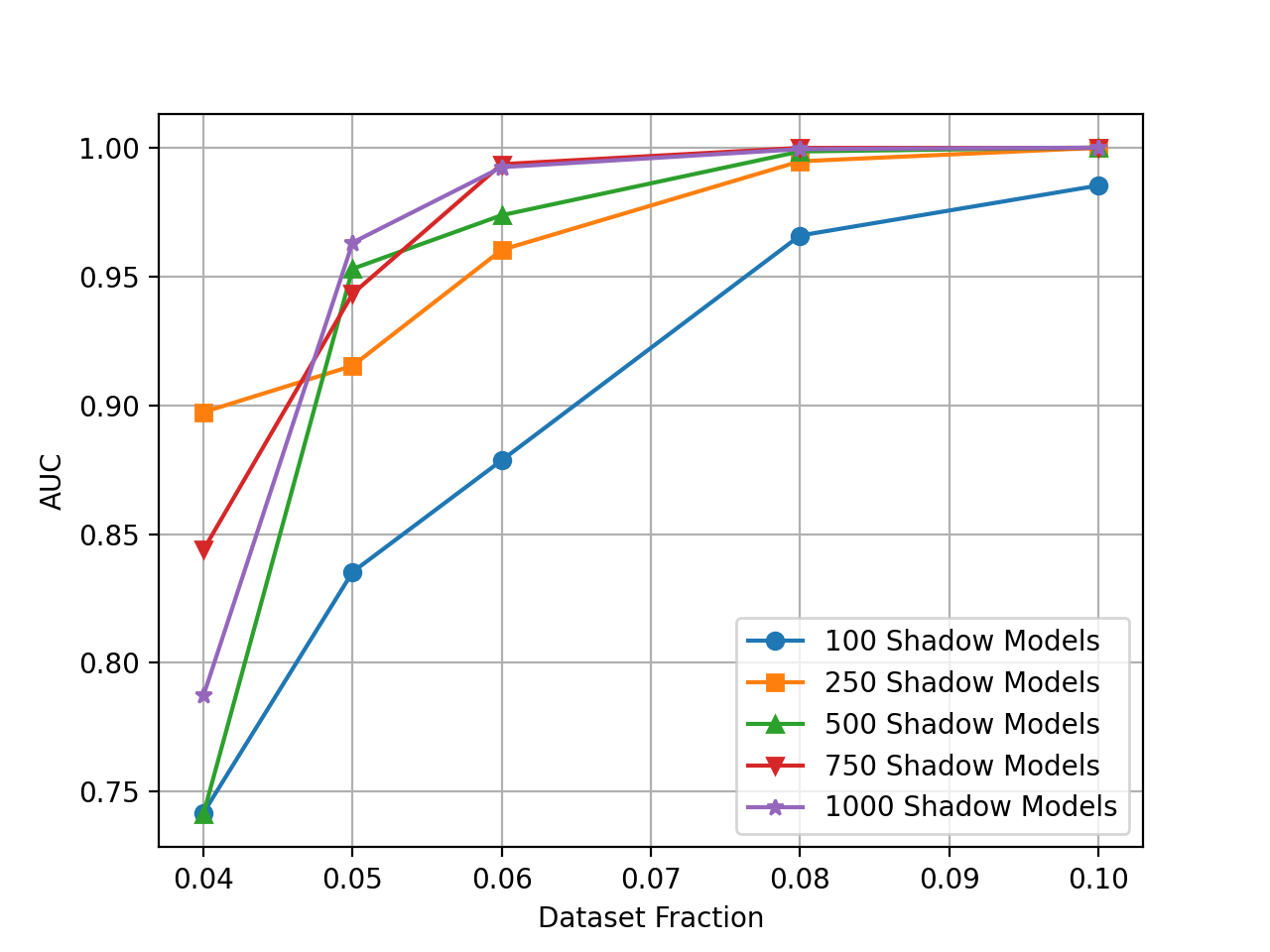}
    \caption{Average AUC with different fractions of the Tiny ImageNet dataset used to train the shadow models across different numbers of shadow models used to train the detector.}
    \label{fig:shadow-models}
\end{figure}

In another experiment, we study the effect of reducing the number of shadow models used to train the detector. We originally trained 1000 models for each fraction of the Tiny ImageNet dataset. We evaluate the perfomrnace when only 100, 250, 500, or 750 models are used to train the detector. The results are shown in \cref{fig:shadow-models}. In these results, the ResNet10 architecture is used for the testing models. Each point in the plot is an average of 10 different runs. The performance does not degrade much if we reduce the number of shadow models down to 250, especially if we use at least 8\% of the Tiny ImageNet dataset for training the shadow models. However, going further down to 100 models can hurt the performance.

\section{Conclusion}
In this paper, we present a new method for detecting Trojan models named TRIGS for TRojan Identification from Gradient-based Signatures. TRIGS applies a data-driven approach, where a signature of a trained model is constructed using activation optimization, and a classifier detects whether the model is Trojan or not based on the signature. On two public datasets as well as our own created challenging dataset, TRIGS achieves state-of-the-art performance, in most cases surpassing baseline methods by large margins. TRIGS works well regardless of whether the probe model architecture is convolutional or a vision transformer. It also works very well when the defender only has access to a small amount of clean samples. Our dataset will be the first public dataset for Trojan detection that is composed only of models based on the vision transformer architecture and trained on a 1000-class classification task (those of the ImageNet dataset).

\section*{Acknowledgement}
This research is based upon work supported by the Defense Advanced Research Projects Agency (DARPA), under cooperative agreement number HR00112020009. The views and conclusions contained herein should not be interpreted as necessarily representing the official policies or endorsements, either expressed or implied, of DARPA or the U.S. Government. The U.S. Government is authorized to reproduce and distribute reprints for governmental purposes notwithstanding any copyright notation thereon.

{\small
\bibliographystyle{ieeenat_fullname}
\bibliography{arxiv24_trojdet}
}

\end{document}